%% file: main.tex
\documentclass[runningheads]{llncs}
\usepackage{graphicx}
\usepackage{amsmath,amssymb} 
\usepackage{color}
\usepackage{wrapfig}
\usepackage{mathtools}
\usepackage[width=122mm,left=12mm,paperwidth=146mm,height=193mm,top=12mm,paperheight=217mm]{geometry}

\DeclareMathOperator*{\argmin}{arg\,min}

\newcommand{\bx}{\mathbf{x}}
\newcommand{\br}{\boldsymbol{\delta}}

\newcommand{\bbx}{\bar{\mathbf{x}}}
\newcommand{\bbz}{\bar{\mathbf{z}}}
\newcommand{\bz}{\mathbf{z}}

\newcommand{\bV}{\mathbf{V}}

\newcommand{\mcF}{\mathcal{F}}
\newcommand{\Exp}[1]{\mathbb{E}\left[#1\right]}
\newcommand{\expon}[1]{\textrm{exp}\left\{#1\right\}}

\begin{document}

\pagestyle{headings}
\mainmatter
\def\ECCV16SubNumber{1706}  

\title{Deep Manifold Traversal: Changing Labels with Convolutional Features} 



\author{Jacob R. Gardner\textsuperscript{*}, Paul Upchurch\textsuperscript{*}, Matt J. Kusner, Yixuan Li, Kilian Q. Weinberger, Kavita Bala, John E. Hopcroft}
\institute{Cornell University, Washington University in St. Louis \\ \scriptsize{*Authors contributing equally}}

\maketitle

\begin{abstract}
Many tasks in computer vision can be cast as a ``label changing''
problem, where the goal is to make a semantic change to the appearance of an image or
some subject in an image in order to alter the class membership.
Although successful task-specific methods have been developed for some label changing applications, to date no general purpose method exists. Motivated by this we propose \emph{deep
manifold traversal}, a method that addresses the problem in its most general form: it first approximates the
manifold of natural images then morphs a test image along a traversal path away from a source class and towards a target class while staying near the manifold throughout. The resulting algorithm is surprisingly effective and versatile. It is completely data driven, requiring only an example set of images from the desired source and target domains.
We demonstrate deep manifold traversal on highly diverse label changing tasks: 
changing an individual's appearance (age and hair color), changing the season of an outdoor image, and transforming a city skyline towards nighttime.
\end{abstract}

\input{intro}
\input{related}

\section{Background: Maximum Mean Discrepancy}
\input{conv}
\input{mmd}
\input{method}
\input{results}
\input{paul}
\input{limitations}

\input{conclusion}

\bibliographystyle{splncs}
\bibliography{iclr2016_conference,matt}
\end{document}

%% file: intro.tex
\section{Introduction}

Many tasks in computer vision can be cast as a \emph{label changing} problem: given an input image, change the label of that image from some label $y^{s}$ to some target label $y^{t}$.
 Recent examples of this general task include changing facial expressions and hairstyle \cite{chai2012single,kemelmacher2013internet}, example-based image colorization \cite{irony2005colorization,gupta2012image}, aging of faces \cite{kemelmacher2014illumination,Boyadzhiev:2015:BDI:2843519.2809796}, material editing~\cite{Boyadzhiev:2015:BDI:2843519.2809796}, editing of outdoor scenes \cite{laffont2014transient} changing seasons \cite{neubert2013appearance}, and image morphs \cite{liao2014automating}, relighting of photos \cite{kopf2008deep,laffont2012coherent} or  hallucinating a night image from a day image \cite{shih2013data}. A variety of specialized algorithms exist for each of these tasks. However, these algorithms often incorporate substantial domain-specific prior knowledge relevant to the task and may require hand annotation of images, rendering them unable to perform any other task. For example, it is unlikely that a facial aging algorithm would be able to change the season of an outdoor scene.

This motivates research into the most \emph{general} form of changing image appearances. Our goal is to design a method that takes as input a set of source and target images (e.g. images of young and old people) and changes a given test image to be semantically more similar to the target  than the source images. 

A given image could be transformed into a target image through linear interpolation in pixel space. However, intermediate images would not be meaningful because the set of natural images does not span a linear subspace in the pixel space. 
Instead, it is believed to constitute a low dimensional sub-manifold~\cite{weinberger2006unsupervised}. In order to make meaningful changes, the image traversal path must be confined to the underlying manifold throughout. 


Bengio et al. 2012 \cite{bengio2012better} hypothesizes that deep convolutional networks linearize the manifold of natural images into a subspace of deep features. 
This suggests that convolutional networks, and in particular the feature space learned by such networks, may be a natural choice for solving the label changing problem. However, recent work \cite{szegedy2013intriguing,nguyen2014deep} has demonstrated that this problem can be surprisingly hard for machine learning algorithms. In the context of object classification through convolutional neural networks, it has been shown possible to change the \emph{prediction} of an image with tiny alterations that can be \emph{imperceptible to humans}. Such changes do not affect the appearance of the image and leave the \emph{class label} untouched~\cite{szegedy2013intriguing}.
In fact, the problem of changing class labels persists for most discriminative machine learning algorithms~\cite{goodfellow2014explaining} and is still an open problem to date. 


In this paper we investigate how to make \emph{meaningful} changes to input images while staying on the underlying manifold. We follow the intuition by Bengio et al. 2012~\cite{bengio2012better} and utilize a deep convolutional network trained on 1.2 million images~\cite{simonyan2014very} to simplify the manifold of natural images to a linear feature space. We avoid the difficulty pointed out by Szegedy et al~\cite{szegedy2013intriguing} by using kernel Maximum Mean Discrepancy (MMD)~\cite{gretton2012kernel} to estimate the distributions of source and target images in this feature space to guide the traversal. The traversal stays on the manifold, because it is confined to the subspace of deep features and is forced by the MMD guide to regions that correspond to likely images. Each point along the path can be mapped back to an image with reverse image reconstruction~\cite{mahendran15understanding}.  Furthermore, our method is linear in space and time so it naturally scales to large images (e.g., 900$\times$600), which is much larger than most results demonstrated by generative models.

In a nutshell, our algorithm works in three steps: 1. Source, target and the test images are forward propagated through a convolutional network and mapped into a deep feature space; 2. MMD is used to guide the traversal of the test image in the deep feature space towards the target and away from the source distribution while staying close to the manifold of natural images; 3. a point along the traversal path is specified and a  corresponding image is generated through reverse image reconstruction. 

The resulting algorithm allows us to traverse the manifold of natural images freely in a completely data-driven way. We only require labeled images from the source and target classes, and no hand annotation (e.g., correspondences or strokes).  
While this method certainly does not replace specialized methods, it may function as a baseline for a wide variety of tasks, or perhaps enable some tasks for which no specialized algorithms have been derived. Our results indicate that our method is highly general and performs better than current general methods (which make use of image morphing) on a number of different tasks.

%% file: related.tex
\section{Related Work}
Szegedy et al.~\cite{szegedy2013intriguing} were the first to show that deep networks can be `easily convinced' that an input is in a different class, by making subtle, imperceptible changes to the input. Such changed inputs were termed `adversarial examples' and Goodfellow, et al.~\cite{goodfellow2014explaining} showed that these examples are generally problematic for high-dimensional linear classifiers. These results indicate it is inherently difficult to meaningfully change the label of an input with small changes. 


In general, generative networks are somewhat orthogonal to our problem setting, as they~\cite{goodfellow2014generative,denton2015deep}, (a) deal primarily with generating novel images rather than changing existing ones, and (b) are typically restricted to very low resolution images, such as 32$\times$32. 

Mahendran and Vedaldi~\cite{mahendran15understanding} recovered visual imagery by inverting
deep convolutional feature representations. Their goal was to reveal
invariance by comparing a reconstructed image to the original
image. Gatys, et al.~\cite{gatys2015neural} demonstrated how to transfer the
artistic style of famous artists to natural images by optimizing for
feature targets during reconstruction. We draw upon these works as
means to demonstrate our framework in the image domain. Yet, rather than
reconstructing imagery or transferring style, we construct new images
which have the qualities of a different class.


A few methods in the machine learning literature also deal with data-driven changes to images. Reed et al.~\cite{reed2014learning,reed2015deep} propose to learn a model to disentangle factors of variation (e.g., identity and viewpoint). In our work, we directly minimize the discrepancy between an image and a target sub-manifold inside the semantic space learned by a convolutional network trained on millions of images. An advantage of our approach is the ability to run on much higher resolution images up to 900x600 in this paper, compared to 48x48 images in~\cite{reed2014learning}.


Analogical reasoning methods~\cite{tenenbaum2000separating,hertzmann2001image,memisevic2007unsupervised,mikolov2013distributed,reed2015deep,sadeghi2015visalogy} solve for D in the expression: A is to B as C is to D. Other methods generate images in a controlled fashion~\cite{dosovitskiy2015learning,kulkarni2015deep}. Our method also has multiple inputs but we do not solve for analogies nor do we learn a disentangled model.

In concept, our work is similar to methods~\cite{suwajanakorn2015makes,garrido2014automatic,thies2015real} which use video or photo collections to capture the personality and character of one person's face and apply it to a different person (a form of puppetry~\cite{sumner2004deformation,weise2009face,kholgade2011content}). This difficult problem requires a complex pipeline to achieve high quality results. For example, Suwajanakorn et al.~\cite{suwajanakorn2015makes} combines several vision methods: fiducial point detection~\cite{xiong2013supervised}, 3D face reconstruction~\cite{suwajanakorn2014total}, optical flow~\cite{kemelmacher2012collection} and texture mapping. Our work is conceptually similar because we also use photo collections to define the source and target. However, we can produce plausible results without any additional machinery and our usage of a high-level semantic CNN feature space makes our method applicable to a wide-variety of domains.

Our task is related to a large body of image morphing work (survey by Wolberg~\cite{wolberg1998image}). Image morphing warps images into an alignment map then color interpolates between mapped points. Unlike image morphing, we do not warp images to a map. A recent work by Liao et al.~\cite{liao2014automating} aligns based on structural similarity~\cite{wang2004image}. Their goal is to achieve semantic alignment partially invariant to lighting, shape and color. We achieve this with a high-level semantic CNN feature space. Their method also requires manual annotations to refine the mapping whereas our method is fully automated.

Kemelmacher et al.~\cite{kemelmacher2011exploring} creates plausible transformations between two images of the same person by selecting an ordered sequence of photos from a large photo collection. Qualitatively, the person may appear to change expression as if the image was changing. Unlike their method, we actually change the original image while preserving the clothing and background.


%% file: conv.tex
%% file: mmd.tex
\label{sec:mmd}

The \emph{Maximum Mean Discrepancy} \cite{fortet1953convergence} (MMD) statistic tests whether two probability distributions, source $P^s$ and target $P^t$, are the same. The MMD metric measures the maximum difference between the mean function values:
\begin{equation}
\text{MMD}(P^s,P^t,\mcF) = \sup_{f \in \mcF} \left(\Exp{f(\bz^s)}_{\bz^s \sim P^s} - \Exp{f(\bz^t)}_{\bz^t \sim P^t}\right)
\end{equation}
given some function class $\mcF$. MMD can be thought of as producing a test function that distinguishes samples from these two distributions. In particular, the MMD test function is large when evaluated on samples drawn from a source distribution $P^s$, and small when evaluated on samples drawn from a target distribution $P^t$. 
When $\mcF$ is a reproducing kernel Hilbert space, the function maximizing this difference can be found analytically \cite{gretton2012kernel}, and is called the \emph{witness function}:
\begin{equation}
	f^{*}(\bz) = \Exp{k(\bz^s,\bz)}_{\bz^s \sim P^s} - \Exp{k(\bz^t,\bz)}_{\bz^t \sim P^t}
\end{equation}
The MMD using this function is a powerful measure of discrepancy between two probability distributions. For example, it is easy to show that if $\mcF$ is universal, then $P^s = P^t$ if and only if $\text{MMD}(P^s,P^t,\mcF)=0$ \cite{gretton2012kernel}. 

Given finite samples $\bz^s_1,...,\bz^s_m \overset{iid}{\sim} P$ and $\bz^t_1,...,\bz^t_n \overset{iid}{\sim} P^t$, the witness function can be estimated empirically:
\begin{equation}
	f^{*}(\bz) \approx \frac{1}{m} \sum_{i=1}^{m} k(\bz^s_{i},\bz) - \frac{1}{n}\sum_{i=1}^{n} k(\bz^t_{i},\bz)
\end{equation}

Intuitively, $f^{*}(\bz)$ measures the degree to which $\bz$ is representative of either $P^s$ --- by taking a positive value --- or $P^t$ --- by taking a negative value. In this work, we will make use of the Gaussian kernel, defined as $k(\bz,\bz')=e^{-\frac{1}{2\sigma}|\bz-\bz'|^2}$, where $\sigma$ is the kernel bandwidth. 
While kernel methods often generalize poorly on images in pixel space because of violated smoothness assumptions, we expect that these assumptions hold after deep visual feature extraction~\cite{bengio2007scaling}. For a more thorough review of the MMD statistic, see \cite{gretton2012kernel}.

\begin{wrapfigure}[27]{R}{0.5\textwidth}
\begin{center}
\centerline{
\includegraphics[width=0.5\textwidth]{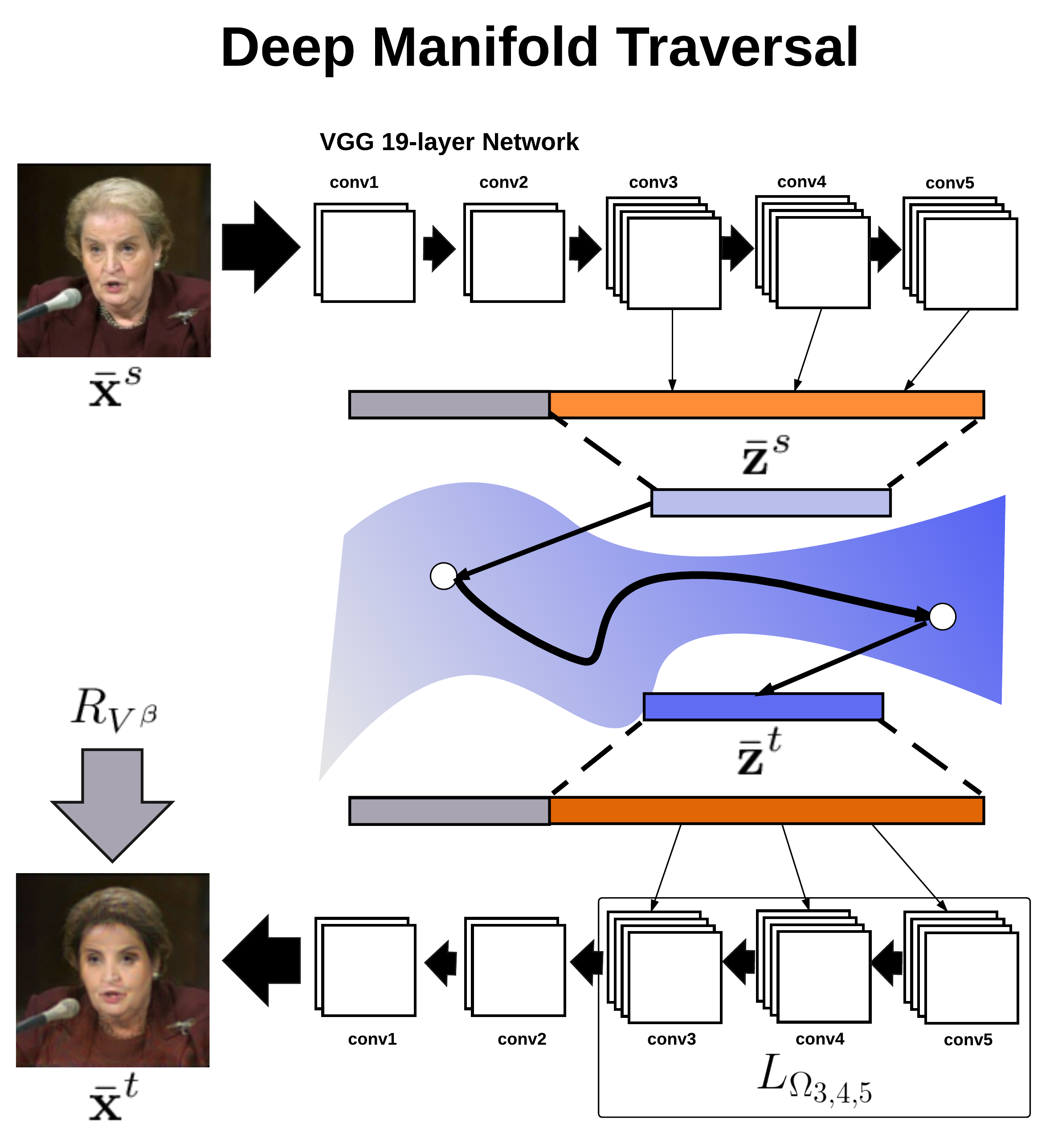}
}
\caption{\textbf{Top:} Input image $\bbx^{s}$ is transformed by a ConvNet to deep features (orange). \textbf{Middle:} The manifold is traversed (black arrow) from source, $\bbz^{s}$, to target, $\bbz^{t}$, in feature space. \textbf{Bottom:} $\bbz^{t}$ is inverted to recover $\bbx^{t}$, subject to total variation regularizer $R_{V^{\beta}}$.}
\label{fig:overview}
\end{center}
\end{wrapfigure}

%% file: method.tex
\section{Deep Manifold Traversal}
\label{sec:method}

In this section, we will discuss our method for manifold traversal from one class into another. Importantly, any transformation should preserve the class-independent aspects of the original image, only changing the class-identifying features. In our setting, we are given a labeled set of images from a \emph{source domain}, $\bx^{s}_{1},...,\bx^{s}_{m}$ each with source label $y^{s}$, and a set of labeled images from a \emph{target domain}, $\bx_{1}^{t},...,\bx^{t}_{n}$ each with target label $y^{t}$. We are also given a specific \emph{input image} $\bbx^{s}$ with label $y^{s}$. Informally, our goal is to change $\bbx^{s} \to \bbx^{t}$ in a meaningful way such that $\bbx^{t}$ has true label $y^{t}$. Figure~\ref{fig:overview} provides an overview of our approach. 

\paragraph{Manifold representation.} The first step of our approach is to approximate the manifold of natural images and obtain a mapping from input images in pixel space, $\bx$, to a high-level feature representation, $\bx_{i} \longrightarrow \phi_{i}$. By modifying these deep visual features rather than the raw pixels of $\bx$ directly, we make changes to the image in a space in which the manifold of natural images is simplified, which more easily allows for images to remain on the manifold.

\paragraph{Network details.} Following the method of~\cite{gatys2015neural} we use the feature representations from deeper layers of a normalized, 19-layer VGG~\cite{simonyan2014very} network. Specifically, we use layers conv3\_1 ($256 \times 63 \times 63$), conv4\_1 ($512 \times 32 \times 32$) and conv5\_1 ($512 \times 16 \times 16$), which have the indicated dimensionalities when the color input is $250 \times 250$. These layers are the first convolutions in the 3rd, 4th and 5th pooling regions. After ReLU, flattening and concatenation, a feature vector has 1.67 million dimensions for a $250 \times 250$ input image.

\paragraph{Image transformation.} Our approach to image transformation will be to change the deep visual features $\bbz^{s}=\phi(\bbx^{s})$ to look more like the deep visual features characteristic of label $y^{t}$. Because the deep convolutional network has mapped the original images in to a more linear subspace, we move linearly away from source high-level features and towards target high-level features. Specifically, we seek to add some linear combination of the source, target, and test images' deep features:

\begin{equation}
\bbz^{t} = \phi(\bbx^{t}) = \bbz^{s} + \bV\br. 
\end{equation}

where $\bV\in\mathbb{R}^{K\times D}$ is the matrix of deep convolutional features for the source, target, and test images: $[\phi_{1}^{t},...,\phi^{t}_{n},\phi^{s}_{1},...,\phi^{s}_{m},\bbz^{s}]$. This linear combination should produce a set of deep features less like the source domain, more like the target domain, but still strongly like the original image. Thus, $\br$ should ideally contain negative values in most source indices, and positive values in most target indices.

To obtain this transformation, we propose an optimization guided by the MMD witness function from section~\ref{sec:mmd}. We make use of the empirical witness function $f^{*}(\bbz^{s}+\bV\br)$ to measure the degree to which the transformed VGG features $\bbz^{s}+\bV\br$ resembles objects with source label $y^{s}$ or those with target label $y^{t}$:
\begin{equation}
f^{*}(\bbz^{s} + \bV\br) = \frac{1}{m} \sum_{i=1}^{m} k(\phi^{s}_{i},\bbz^{s}+\bV\br) - \frac{1}{n} \sum_{j=1}^{n} k(\phi^{t}_{j},\bbz^{s}+\bV\br).
\end{equation}
Observing that--given the definition of $\bV$--each $\phi^{s}_{i}$ and $\phi^{t}_{j}$ and $\bbz^{s}$ can be themselves written as $\bV e_{i}$, $\bV e_{j}$ and $\bV e_{K}$ for one-hot vectors $e_{i}$, $e_{j}$, and $e_{K}$ we rewrite the above as:
\begin{equation}
f^{*}(\bV e_{K} + \bV\br) = \frac{1}{m} \sum_{i=1}^{m} k(\bV e^{s}_{i},\bV e_{K} + \bV\br) - \frac{1}{n} \sum_{j=1}^{n} k(\bV e^{t}_{j},\bV e_{K} + \bV\br).
\end{equation}
When using the squared exponential kernel, we can factor $\bV$:
\begin{align}
f^{*}(\bV e_{K} + \bV\br) & = \frac{1}{m} \sum_{i=1}^{m} \expon{-\frac{1}{\sigma}(e^{s}_{i} - (e_{K} + \br))\bV^{\top}{\bV}(e^{s}_{i}-(e_{K} + \br))}\nonumber \\
                 & - \frac{1}{n} \sum_{j=1}^{n} \expon{-\frac{1}{\sigma}(e^{t}_{j} - (e_{K} + \br))\bV^{\top}{\bV}(e^{t}_{j}-(e_{K} + \br))}.
\end{align}
If the $K\times K$ matrix $\bV^{\top}\bV$ is precomputed for a dataset, this function can be computed in time \emph{independent} of the number of convolutional features, and therefore original image resolution. 

The witness function $f^{*}(\bV e_{K} + \bV\delta)$ has a negative value if the transformed visual features $\bV e_{K} + \bV\delta$ are more characteristic of label $y^{t}$ than of label $y^{s}$. To transform $\bbx^{s}$ to have target label $y^{t}$, we therefore wish to minimize $f^{*}(\bV\phi(\bbz)^{s}+\bV\br)$ in $\br$. However, when performed unbounded, this optimization moves too far along the manifold to a mode of the target domain, preserving little of the information contained in $\bbz^{s}$. We therefore follow the techniques used in~\cite{szegedy2013intriguing} and enforce a \emph{budget} of change, and instead obtain $\bbz^{t}$ by minimizing:
\begin{equation}
\phi(\bbz^{t}) = \bV(e_{K} + \br) \ \ \  \textrm{ where: }\  \br=\argmin_{\br} f^{*}(\bV e_{K} + \bV\br) + \lambda \Vert \bV\br \Vert^{2}_{2}
\end{equation}
Minimizing the witness function encodes two ``forces'': $\phi(\bbz^s)$ is pushed \emph{away} from visual features characteristic of the source label $y^{s}$ and simultaneously pulled \emph{towards} visual features characteristic of the target label $y^{t}$. 
\paragraph{Reconstruction.}
The optimization results in the transformed representation $\bbz^{t}=\bV e_{K} + \bV\br$. In order to obtain our corresponding target image $\bbx^t=\phi^{-1}(\bbz^t)$, we need to ``invert'' the CNN.
The deep CNN mapping is not invertible, so we cannot obtain the image in pixel space $\bbx^t$  from $\bbz^t$ directly. The mapping is however differentiable and we can adopt the approaches of~\cite{mahendran15understanding} and~\cite{gatys2015neural} to find $\bbx^t$ with gradient descent by minimizing the loss function
\begin{equation}
L_{\Omega_{3,4,5}}(\bbx^{t}) = \frac{1}{2} \Vert\Omega_{3,4,5}(\bbx^{t}) - \bbz^{t}\Vert^2.\label{eq:under}
\end{equation}


\paragraph{Regularization.}


Following the method of~\cite{mahendran15understanding}, we add a total variation regularizer
\begin{equation}
R_{V^{\beta}}(\bbx^{t}) = \sum_{i,j} \left( (x_{i,j+1} - x_{i,j})^2 + (x_{i+1,j} - x_{i,j})^2 \right)^{\frac{\beta}{2}}.
\end{equation}
Here, $x_{i,j}$ refers to the pixel with $i,j$ coordinate in image $\bx$. 
The addition of this regularizer greatly improves image quality. The final optimization problem becomes
\begin{equation}
\bbx^t=\argmin_{\bbx^t} L_{\Omega_{3,4,5}}(\bbx^{t}) + \lambda_{V^{\beta}} R_{V^{\beta}}(\bbx^{t}).\label{eq:finalobj}
\end{equation}
We minimize (\ref{eq:finalobj}) with bounded L-BFGS initialized with $\br=0$. We set $\lambda_{V^{\beta}} = 0.001$ and $\beta = 2$ in our experiments. After reconstruction we have completed the manifold traversal from source to the target: $\bbx^{s}\!\to\! \bbz^{s}\! \to\! \bbz^{t}\! \to\! \bbx^{t}$. We will provide source code for our method on GitHub at http://anonymized.

%% file: results.tex
\section{Experimental Results - LFW}


We evaluate our method on several manifold traversal tasks using the Labeled Faces in the Wild (LFW) dataset. This dataset contains 13,143 images (250$\times$250) of faces with predicted annotations for 73 different attributes (e.g., ``sunglasses'', ``soft lighting'', ``round face'', ``curly hair'', ``mustache'', etc.). We use these annotations as labels for our manifold traversal experiments. Because the predicted annotations \cite{CAVE_0296} have label noise, we take the 2,000 most confidently labeled images to construct an image set. For example, in our aging task below, we take the bottom (i.e., most negative) and top (i.e., most positive) 2,000 images in the ``senior'' class as our source and target image sets. 

All single transformation image results shown for LFW use the same $\lambda$ value of $\lambda=\textrm{4e-8}$. All experiments were run with RBF kernel width $\sigma=\textrm{7.7e5}$. In the tasks below, test images were chosen at random, with the exception of Aaron Eckhart (the first image in LFW), who we included in all tests in order to show multiple tasks on the same image. Due to space constraints we only show a small number of results per experiment. More results are in the supplemental.
\input{aging_text}
\input{face_baselines.tex}

\input{adversarial_text}
\input{hair_color_text}

\section{Experimental Results - AMOS}
Does our technique work outside the context of faces? To test this, we also evaluate our method on two tasks using data from the Archive of Many Outdoor Scenes (AMOS) collection of webcams \cite{jacobs2007consistent}. This dataset contains images from thousands of webcams taken nearly hourly (with some missing data) over the course of several years. While this data lacks the rich set of annotations that LFW has, we are able to construct two tasks based on image timestamps--traversing from winter to summer and traversing from day to dusk.
\input{seasons_text}
\input{daynight_text}

%% file: aging_text.tex
\subsection{Aging faces via manifold traversal.}
\label{sec:aging}
To demonstrate the ability of our algorithm to make meaningful changes to the true label of a test instance, we first consider the task of computationally aging faces. To do this, we first follow our procedure above for selecting 2,000 source (young) and target (old) images. We select 7 test images at random plus Aaron Eckhart from the remainder of LFW. We then perform manifold traversal towards ``senior'' on these 8 images, using the same value of $\lambda$ for each traversal. 

\begin{figure*}[tb]
\begin{center}
\centerline{\includegraphics[width=\textwidth]{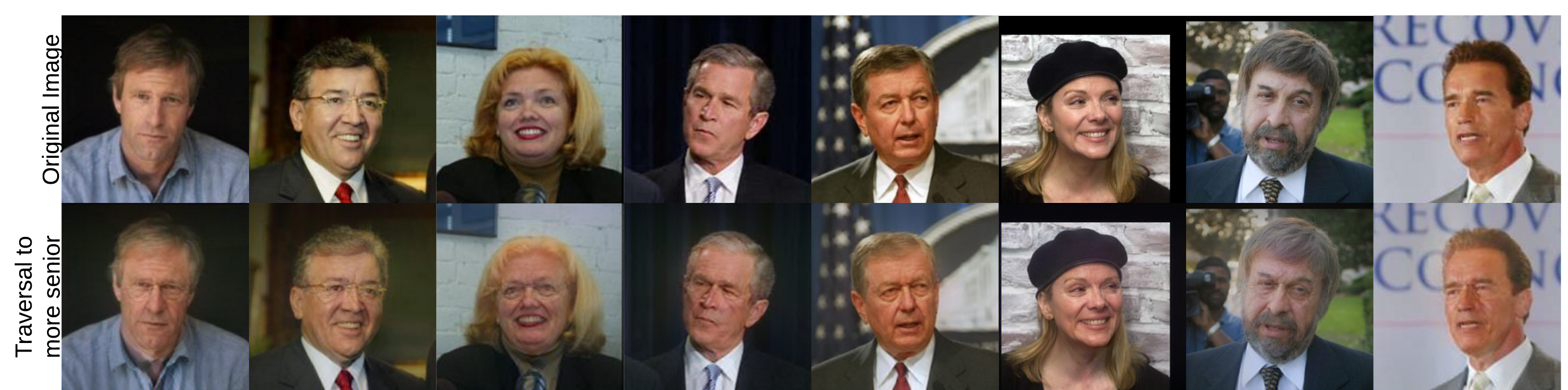}}
\vspace{-2ex}
\caption{\textbf{(Zoom in for details.)} Face aging via manifold traversal on random (except Aaron Eckhart) 250x250 test images from LFW. All aging results shown were run with the same value of $\lambda$.}
\vspace{-7ex}
\label{fig:aging}
\end{center}
\end{figure*}

The results of our aging experiment are shown in figure \ref{fig:aging}. In each case, deep manifold traversal generates an older-appearing version of the original image by adding wrinkles, graying hair and adding bags under eyes. Note that the images remain sharp despite the relatively high resolution compared to existing purely learning-based approaches for facial morphing \cite{reed2014learning}.

One important aspect of the transformations made by deep manifold traversal is that changes are localized to the face and hair. Clothing, background, lighting, and other features of the image irrelevant to the desired label change were not significantly affected. Thus, our algorithm succeeds in preserving as much character of the original image as possible while still changing the true label of the image.

Finally, we note that an advantage of our technique over many other approaches is that we do not need a photocollection of the test individual. For example, Aaron Eckhart and Mark Rosenbaum (first and 7th column in the figure) only have one image in the dataset. 

%% file: face_baselines.tex
\paragraph{Comparison on aging.}

In this section, we compare several methods to deep manifold traversal on the aging task. We compare to two alternative data driven approaches that motivate the need for performing traversal with deep features. First, we compare to ``shallow'' manifold traversal, where we perform our linear traversal algorithm, but in the original pixel space rather than after extracting deep convolutional features. We also compare to interpolation in pixel space between the original input image and the average ``senior'' image. We also compare to a state-of-the-art technique for image morphing \cite{liao2014automating}, which only requires one target image but requires manual annotation of correspondances between the test and target image.

In the case of aging, the image morphing algorithm requires both a young and an aged photo of the same person, which would not typically be available. Therefore, we chose to evaluate the aging task on Harrison Ford, as young and old images of him are both readily available from Google image search. For the image morphing baseline, we show the ``halfway'' image. The annotationed correspondences are shown as red dots on the original and target image.

\begin{figure*}[tb]
\begin{center}
\centerline{\includegraphics[width=\textwidth]{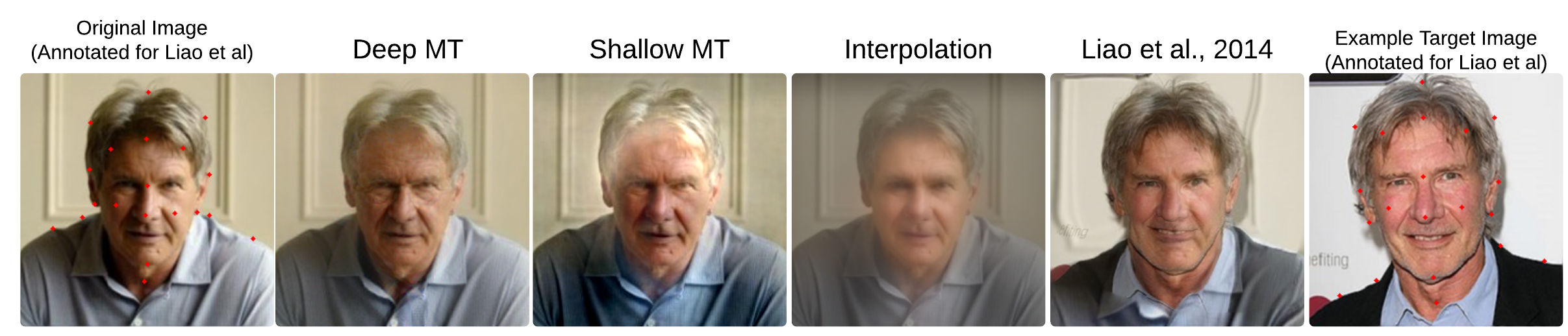}}
\vspace{-2ex}
\caption{\textbf{(Zoom in for details.)} Several methods used to change the age of an input image of Harrison Ford.}
\vspace{-7ex}
\label{fig:ID}
\end{center}
\end{figure*}

The results of our experiment are shown in figure \ref{fig:ID}. Deep manifold traversal clearly perorms better than both of the other data-driven baselines, producing a sharp image with characteristic aging features. This suggests that traversal in the deep convolutional feature space is indeed necessary. When compared to the image morphing task, the visual clarity of the face are comparable. However, the image morphing algorithm introduces some warping of the face in the intermediate stages. Perhaps the most obvious difference between the two methods is that deep manifold traversal preserves both the background and the clothing of the original image, thus avoiding changes that are irrelevant to the desired change.

%% file: adversarial_text.tex

\paragraph{Comparison with Szegedy et al. 2014.}

Existing work has shown that it is possible to make imperceptible changes to images so that deep convolutional networks make high-confidence misclassifications \cite{szegedy2013intriguing}. In this section, we demonstrate that when we vary $\lambda$ in our manifold traversal algorithm, we can gradually change both the class label of an image \emph{and} a machine learning classifier's prediction, not just the prediction alone.

To do this, we use the convolutional layers of VGG as a feature extractor, and train an SVM using the top 2000 ``senior'' and ``non-senior'' faces from LFW to distinguish between VGG features extracted from images with positive ``senior'' attribute values and negative ones. We then use Platt scaling to transform the SVM decision values into probabilities \cite{platt1999probabilistic} ranging between $0$ and $1$, where lower probability value indicates the likelihood for being more ``senior''. 

\begin{figure*}[tb]
\begin{center}
\centerline{\includegraphics[width=\textwidth]{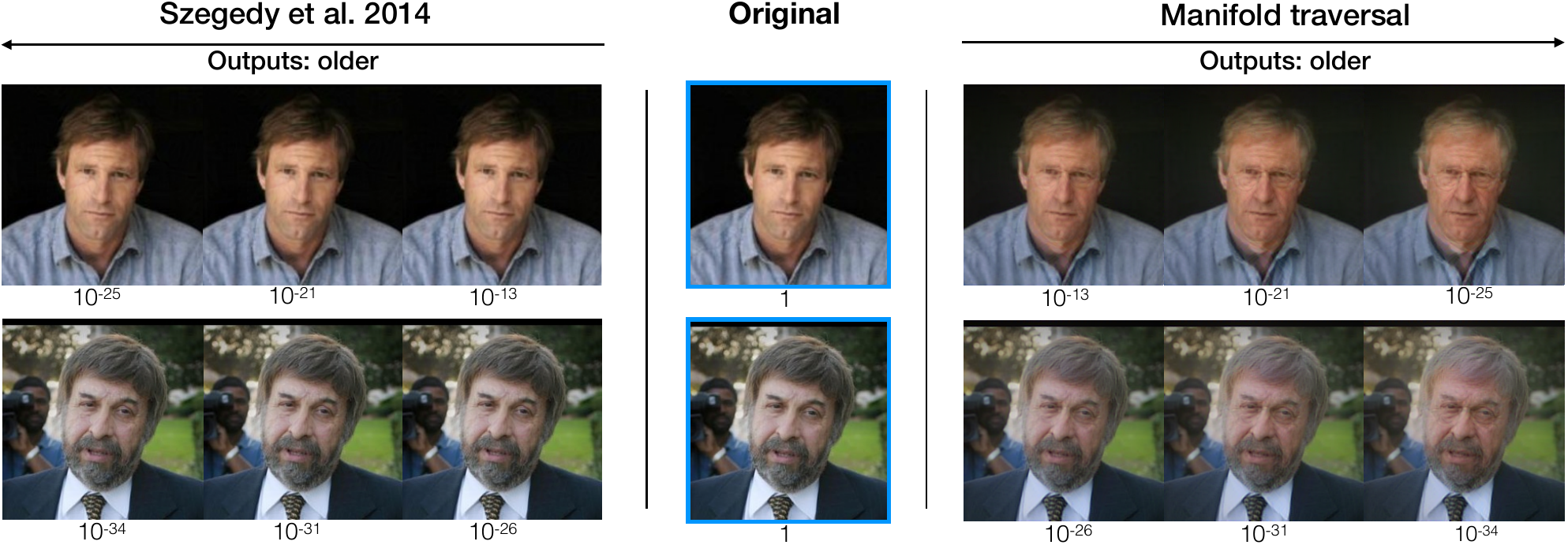}}
\vspace{-3ex}
\caption{\textbf{(Zoom in for details.)} \textbf{Left:} ``Aging'' images generated using the method of \cite{szegedy2013intriguing}. \textbf{Right:} ``Aging'' images generated by deep manifold traversal. The image progression towards the right was generated by gradually decreasing the value of $\lambda$. Numbers below each image show the Platt scaled probabilities of an SVM trained on VGG features to distinguish old age, where lower values indicate more ``senior''.}
\vspace{-9ex}
\label{fig:advers}
\end{center}
\end{figure*}

We construct adversarial ``senior'' images--which we display on the left in figure \ref{fig:advers}--as well as perform manifold traversal with three different lambdas, which we display on the right in figure \ref{fig:advers}. All manifold traversal results were generated using the same set of lambda values: 6e-8, 5e-8, and 4e-8. 

Below each image is the class probability of ``not senior'' assigned to that image by the Platt-scaled SVM. In order to make outputs on both sides comparable, we set the adversarial regularizer so that the adversarial images have comparable decision values to those generated by manifold traversal.  

We note several important features of this result. First, the original images all have very high probability of being ``not senior''. However, after both the adversarial and the DMT modifications, we were able to change the SVM prediction to be completely confident that the transformed images were of seniors. We find that deep manifold traversal makes meaningful change to the \emph{true label} of the images as well, clearly aging the person in each image. In contrast, the comparable adversarial images fail to change the original images in a human-perceptibly meaningful way.

%% file: hair_color_text.tex
\subsection{Changing hair color via manifold traversal.}

To show the versatility of manifold traversal, we also perform manifold traversal to change hair color. This task is different from aging because different hair styles require manifold traversal to focus on a larger variety of shapes than aging does. We perform two traversals: one towards blonde (lighter) hair, and one towards black (darker) hair. To help ensure that the randomly selected test images did not already have blonde or black hair, we selected our 8 random test images from among the top 90th percentile of the ``brown hair'' attribute.

\begin{figure*}[tb]
\begin{center}
\centerline{\includegraphics[width=\textwidth]{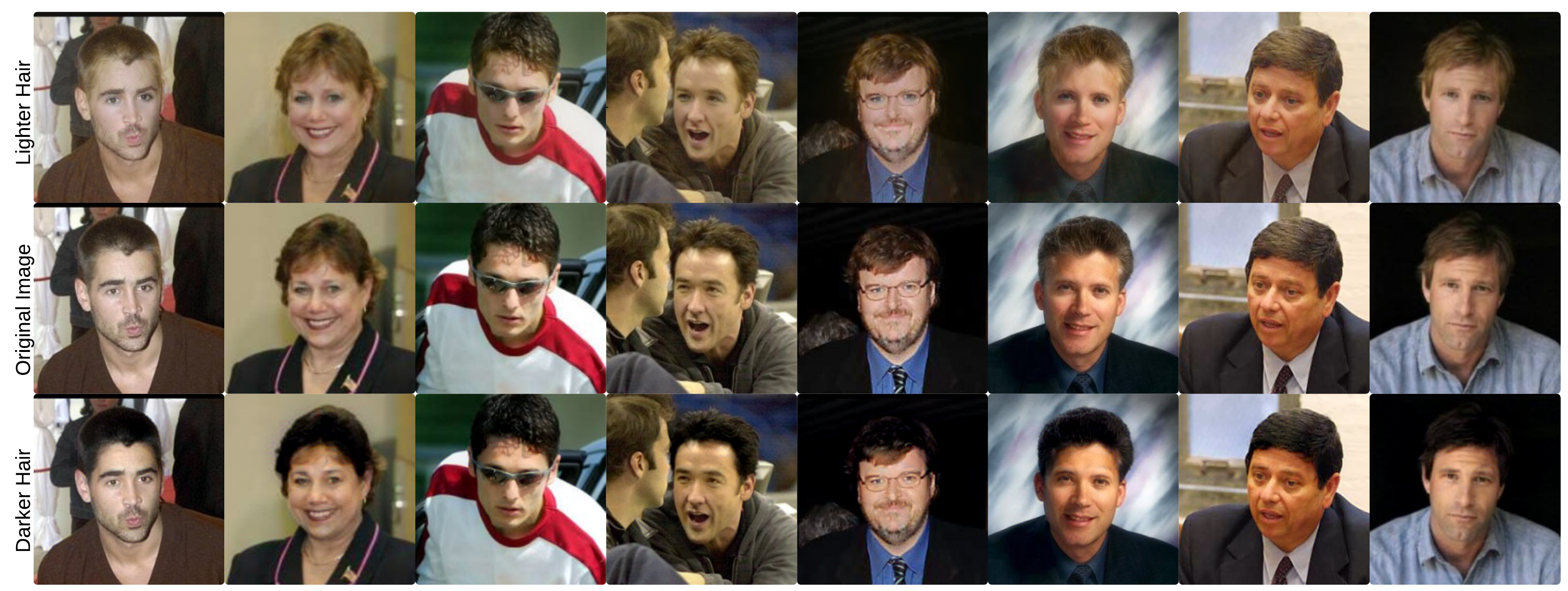}}
\vspace{-2ex}
\caption{\textbf{(Zoom in for details.)} Changing hair color of random (except Aaron Eckhart) 250x250 images from LFW with manifold traversal. \textbf{Top.} Manifold traversal to lighter hair. \textbf{Middle.} Original image. \textbf{Bottom.} Manifold traversal to darker hair. All traversals were performed with the same value of lambda.}
\vspace{-7ex}
\label{fig:hair}
\end{center}
\end{figure*}

The results of our hair color experiment are shown in figure \ref{fig:hair}. The middle row displays the original images in LFW. The top and bottom row show the results of manifold traversal towards lighter hair (``blonde hair'') and darker hair (``black hair'') respectively.

We note that the hair color traversal generally succeeded despite the variety of hair styles, while again preserving features of the image like clothing and background. The varying hair styles suggest that manifold traversal is able to transform more complex shapes than simply faces.

Of particular interest in this experiment are the changes made other than the color of hair on the top of the head. In most cases facial hair such as eyebrows and beard hair was changed to the appropriate color as well (for example in the first column). Furthermore, when traversing to blonde hair, eye color was occasionally also changed to blue to match (for example, in the 2nd, 5th, and 6th columns).

%% file: seasons_text.tex
\subsection{Changing from winter to summer.}

In this section, we look at if we can learn to transform images from winter to summer given a specific webcam. We collect 2762 images from January and February to form the source ``winter'' set, and 2858 images from June and July form the target ``summer'' set. We then select two winter test images which do not occur in either the source or target set and perform deep manifold traversal.

\begin{figure*}[tb]
\begin{center}
\centerline{\includegraphics[width=\textwidth]{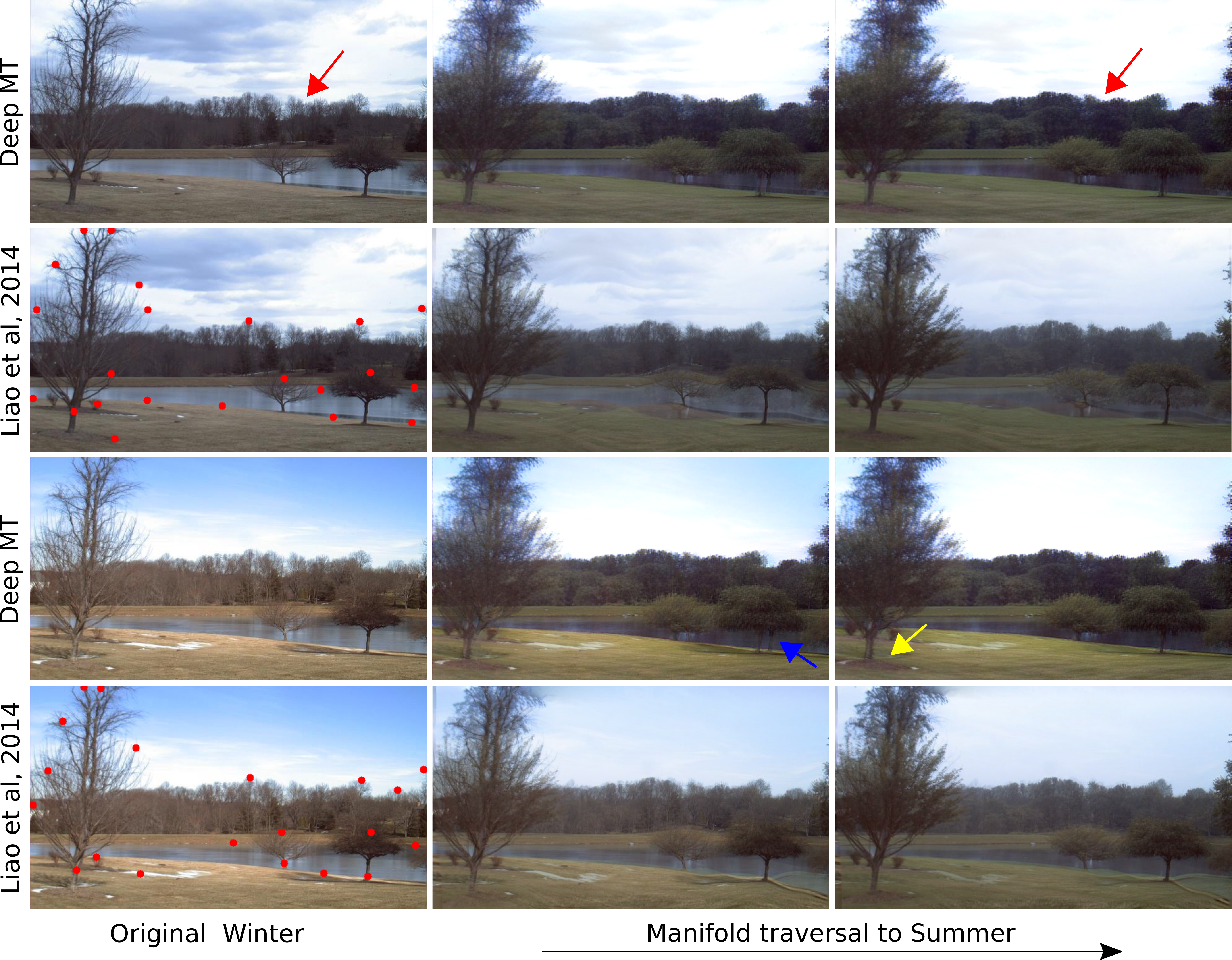}}
\vspace{-2ex}
\caption{\textbf{(Zoom in for details.)} Changing from winter to summer with deep manifold traversal (1st and 3rd row). Tree branches are replaced with leaves (red arrows), dirt appears at the base of a large tree (yellow arrow). At a partial traversal a tree trunk (blue arrow) is duplicated. This may be due to a viewpoint change. For comparison we show image morphing~\cite{liao2014automating} (2nd and 4th row). \cite{liao2014automating} requires manual annotations (red dots) and uses a single target image rather than a photo collection.}
\vspace{-7ex}
\label{fig:seasons}
\end{center}
\end{figure*}

The results of both deep manifold traversal and the image morphing algorithm of \cite{liao2014automating} on this task are shown in figure \ref{fig:seasons}. In both test images, deep manifold traversal adds leaves to the trees in the foreground, as well as dense foliage to the forest in the background. In the second test image, the grass is made significantly greener, and the snow on the ground begins to fade. We notice that during partial traversal a tree trunk is added in the second experiment (likely due to a viewpoint change), which fades upon complete traversal.

The image morphing algorithm also performs reasonably well when adding leaves to trees, producing leaves of comparable quality to deep manifold traversal. However, we note two notable image artifacts in the morphing algorithm results. First, the trunk of the foreground tree is clearly still visible, despite the fact that there is dense foliage. Second, while the image morphing algorithm did not duplicate the trunk of the tree on the right, there is significant image warping near that tree and the bank of the lake. One possible reason for this may be due to the fact that the image morphing technique relies on a single target image. This means that, if a natural event causes the camera viewpoint to change slightly, the algorithm must also morph the viewpoint, which may be the cause of the odd riverbank location in the morphed image. Deep manifold traversal, however, is robust to such changes during full traversal as such small irregularities are not vital to the label change.

%% file: daynight_text.tex
\subsection{Scalability}
\label{sec:daynight}

\begin{figure*}[tb]
\begin{center}
\centerline{\includegraphics[width=\textwidth]{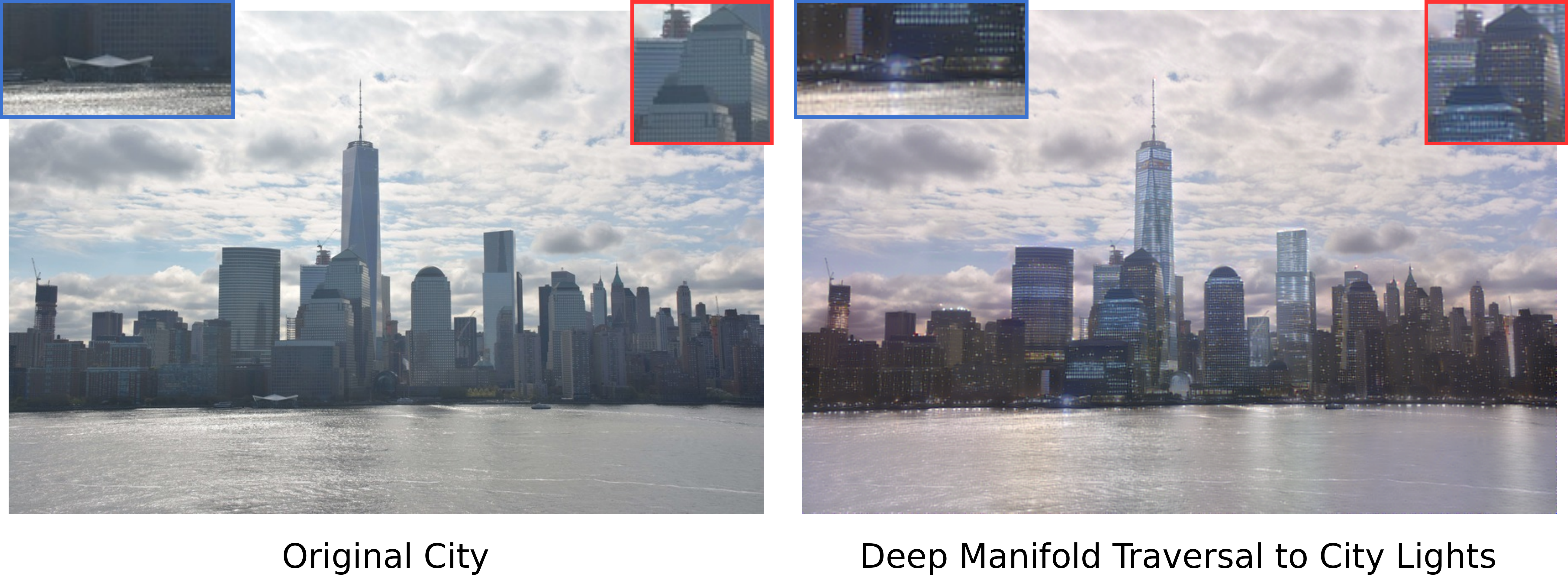}}
\vspace{-2ex}
\caption{\textbf{(Zoom in for details.)} Deep manifold traversal at 900$\times$600 pixels. The city (left) is changed to make it more similar to nighttime (right). Our data-driven method selects multiple factors to change. The tone of the buildings changes from daytime gray to nighttime blue and nighttime artificial lighting appears in windows (red insets). The waterfront pavilion light and car headlamps are reflected on the water (blue insets).}
\vspace{-7ex}
\label{fig:daynight}
\end{center}
\end{figure*}

How well does our method scale to larger images? As a demonstration, we performed a manifold traversal on a 4k resolution AMOS webcam. The images were downsampled to 900$\times$600 then a manifold traversal was performed on a random test image. The traversal was from 2051 day images toward 1507 night images --- day and night selected by timestamp, dawn and dusk excluded. The test image was not one of the source or target images.
Figure~\ref{fig:daynight} shows the traversal result. Our method found that changing tone, adding artificial lighting and reflections of light off the water (see insets in the figure) are the cues which make the image more like nighttime. Interestingly, the sky remains blue as it would during the day. One hypothesis for this is that, because VGG was trained on an object recognition dataset, the sky is treated as background and not represented in the high-level feature space--for example, when classifying birds or airplanes, the sky is background. 

The feature matrix is 3559$\times$14088192, which requires 186 GB of storage. Manifold traversal takes 132 minutes and reconstruction takes 43 minutes. In comparison, LFW (250$\times$250, 2000 source and 2000 target images) requires 25 GB (feature matrix is 4001$\times$1671424) and 18 minutes to transform. Our method can transform large images (larger than most generative model demonstrations) and is primarily limited by memory constraints. Furthermore, the manifold traversal time is linear in image size.

%% file: paul.tex

%% file: limitations.tex
\section{Discussion and Future Work}

\begin{wrapfigure}[9]{R}{0.40\textwidth}
\vspace{-5ex}
\begin{center}
\includegraphics[width=0.39\textwidth]{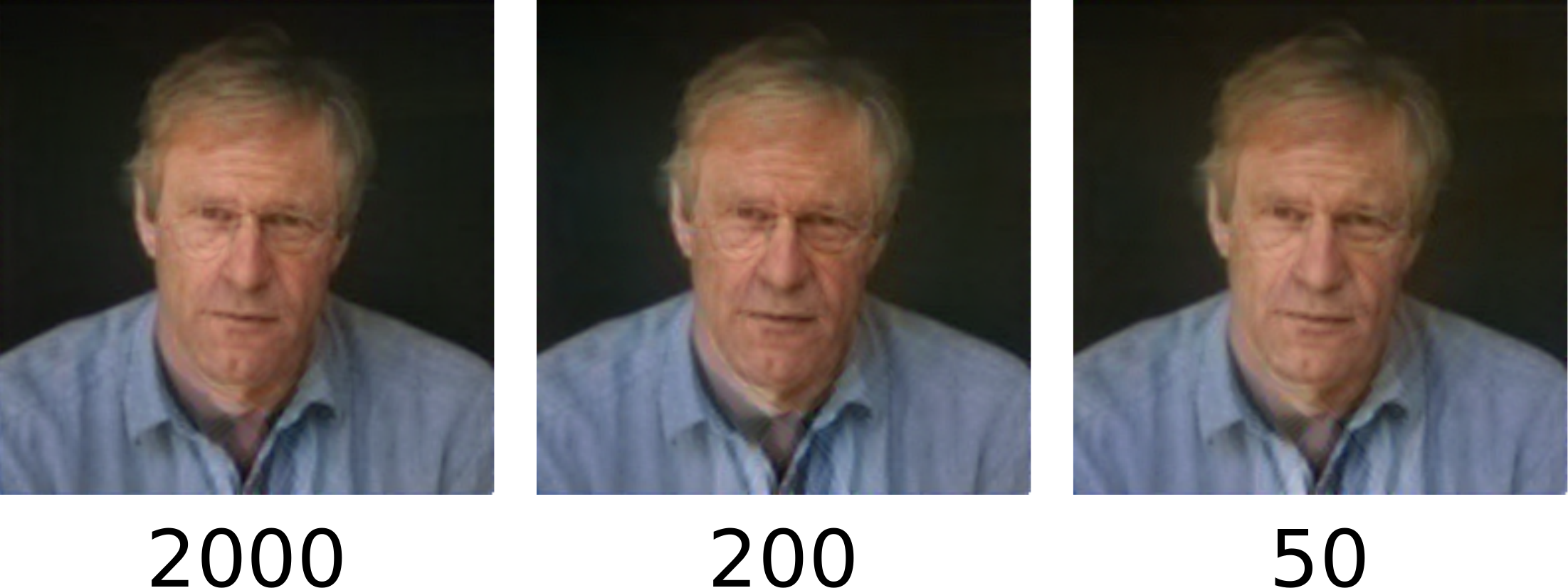}
\end{center}
\vspace{-4ex}
\caption{The effect of varying the number images used to define a manifold. }
\label{fig:varyk}
\end{wrapfigure}

In the LFW experiments we use 2000 source and 2000 target images to define the manifold. It is possible to use fewer images at the cost of reduced output quality (figure~\ref{fig:varyk}). There are ways to address this limitation. Video sources can generally produce thousands of images easily. Data augmenation could increase the effective size of a small image set. Exploring ways to reduce the number of images while maintaining quality would be future research.

We find that images must be well aligned. For example, in figure~\ref{fig:seasons} the small tree on the right is displaced between the source and target image sets. As a result, a ghostly tree trunk appears at some lambdas (but disappears when lambda is sufficiently small). We note that only the subject needs to be aligned. For example, there is variety in the LFW backgrounds yet this does not prevent our method from operating on the aligned faces. It may be possible to overcome this limitation by incorporating an image alignment mapping~\cite{suwajanakorn2015makes} or to automatically identify photos taken from the same viewpoint~\cite{snavely2008modeling}.

Although we gain much from using VGG features, those features are roughly 10x larger than the input image. As a result, holding thousands of 960x540 image feature vectors requires over 128 GB of main memory. These limitations can be overcome by out-of-core methods at the cost of speed. Reducing the size of the deep neural network feature space is future research.

Many of the best state-of-the-art methods are computational pipelines which combine domain-specific knowledge and specialized algorithms to solve sub-problems of a larger problem. An exciting direction of future research is to see if our generic method can simplify existing state-of-the-art methods by replacing pieces of the pipeline with our data-driven approach.

One possible use case for deep manifold traversal is in data augmentation. Typical data augmentation involves transforming images with label-invariant changes such as horizontal flipping, with the goal of constructing a larger dataset. If we seek to train a deep neural network that, for examples, distinguishes between young and old faces, we could augment our data by performing manifold traversal on other aspects--such as facial expressions or hair color.

\section{Conclusion}
\label{sec:conclusion}

We introduced a single general purpose approach to make semantically meaningful changes to images in an automated fashion. 
In contrast to prior work, our approach is not specific for any given task. We leverage the combination of MMD and  deep features from convolutional networks to naturally confine the traversal path onto the manifold of natural images. The resulting algorithm scales linearly in space and time (after pre-processing), is extremely general and only requires minimal supervision through example images from source and target domains. However, we believe that the true power of our method lies in its versatility. 
Without modifications it can be applied to changing the appearance of faces, city skylines or nature scenes. 
As future work we plan to investigate the use of manifold traversal for active learning and automated image augmentation as pre-processing for supervised computer vision tasks. We hope that our work will be used as a baseline for a variety of computer vision tasks and will enable new application in areas where no specialized algorithms exist.